\documentclass{article}

\usepackage{arxiv}

\usepackage[utf8]{inputenc} 
\usepackage[T1]{fontenc}    
\usepackage{hyperref}       
\usepackage{url}            
\usepackage{booktabs}       
\usepackage{amsfonts}       
\usepackage{nicefrac}       
\usepackage{microtype}      
\usepackage{lipsum}		
\usepackage{graphicx}
\usepackage{amsmath}
\usepackage{doi}

\title{REG: Refined Generalized Focal Loss for Road Asset Detection on Thai Highways Using Vision-Based Detection and Segmentation Models}

\author{ \href{https://orcid.org/0000-0001-8464-4476}{\includegraphics[scale=0.06]{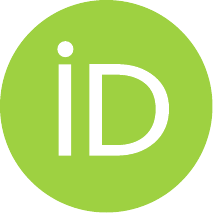}\hspace{1mm}Teerapong Panboonyuen}\thanks{I thank myself for making this work possible, hoping it helps improve vision-based models and inspires others. Explore more about me at \href{https://kaopanboonyuen.github.io/}{https://kaopanboonyuen.github.io/}.} \\
	Postdoctoral Researcher, Chulalongkorn University\\
	Senior Research Scientist, MARS (Motor AI Recognition Solution)\\
	\texttt{teerapong.panboonyuen@gmail.com} \\
}

\renewcommand{\shorttitle}{Teerapong Panboonyuen}

\hypersetup{
pdftitle={REG},
pdfsubject={AI},
pdfauthor={Teerapong Panboonyuen},
pdfkeywords={Computer Vision, AI, Deep Learning},
}

\begin{document}
\maketitle

\begin{abstract}
This paper introduces a novel framework for detecting and segmenting critical road assets on Thai highways using an advanced \textbf{Re}fined \textbf{G}eneralized Focal Loss (REG) formulation. Integrated into state-of-the-art vision-based detection and segmentation models, the proposed method effectively addresses class imbalance and the challenges of localizing small, underrepresented road elements, including pavilions, pedestrian bridges, information signs, single-arm poles, bus stops, warning signs, and concrete guardrails. To improve both detection and segmentation accuracy, a multi-task learning strategy is adopted, optimizing REG across multiple tasks. REG is further enhanced by incorporating a spatial-contextual adjustment term, which accounts for the spatial distribution of road assets, and a probabilistic refinement, which captures prediction uncertainty in complex environments, such as varying lighting conditions and cluttered backgrounds. Our rigorous mathematical formulation demonstrates that REG minimizes localization and classification errors by applying adaptive weighting to hard-to-detect instances while down-weighting easier examples. Experimental results show a substantial performance improvement, achieving a mAP50 of 80.34 and an F1-score of 77.87, significantly outperforming conventional methods. This research underscores the capability of advanced loss function refinements to enhance the robustness and accuracy of road asset detection and segmentation, thereby contributing to improved road safety and infrastructure management. For an in-depth discussion of the mathematical background and related methods, please refer to previous work available at \url{https://github.com/kaopanboonyuen/REG}.
\end{abstract}

\section{Introduction}

The task of detecting and segmenting road assets, such as pavilions, pedestrian bridges, and information signs, is pivotal for modern infrastructure management and road safety. Accurate identification of these assets is crucial for preventing accidents and maintaining road systems efficiently. While vision-based detection REGs have achieved substantial success across various domains \cite{he2017mask, lin2017focal}, these REGs often encounter difficulties when applied to real-world conditions characterized by class imbalance and complex backgrounds \cite{zhang2020bridging, panboonyuen2019semantic, panboonyuen2023mars}. For road asset detection, small, underrepresented classes, such as single-arm poles and bus stops, pose significant challenges in localization and segmentation.

Recent advancements in deep learning REGs for object detection and segmentation have led to increasingly sophisticated architectures, including Transformer-based REGs \cite{carion2020end}, attention mechanisms \cite{vaswani2017attention}, and feature pyramids \cite{lin2017feature}. Despite these advancements, these REGs often experience performance degradation when tasked with detecting small objects in highly cluttered environments \cite{liu2018path}. To address these limitations, we propose a refined version of the Generalized Focal Loss (GFL), which we term Refined Generalized Focal Loss (REG). This refinement builds upon the foundational principles of Focal Loss \cite{lin2017focal}, while introducing modifications to better accommodate spatial context and class uncertainty.

Our contributions are twofold. First, we introduce REG, a new loss function designed to address class imbalance by dynamically adjusting the importance of difficult examples based on spatial context and class rarity. This is achieved through a sophisticated mathematical formulation that incorporates a refinement term \( g_{i,c} \), which accounts for the spatial and contextual significance of each class. The refinement term adjusts the loss according to geometric and contextual factors, providing a more nuanced approach to handling class imbalance and enhancing detection accuracy. The mathematical formulation of REG includes a spatial-contextual adjustment term and a probabilistic refinement that incorporates prediction uncertainty, thus improving model robustness in varied conditions.

Second, we integrate REG within a multi-task framework for simultaneous detection and segmentation, ensuring that the REG learns complementary representations for both tasks \cite{kirillov2019panoptic}. We demonstrate that this approach significantly enhances model robustness and detection accuracy in challenging road environments, showcasing the effectiveness of advanced mathematical techniques in refining loss functions to tackle complex real-world problems.

In summary, the mathematical innovations embedded in the Refined Generalized Focal Loss (REG) provide key insights into its effectiveness. By incorporating the refinement term \( g_{i,c} \), which captures spatial and contextual factors, the loss function dynamically adapts to variations in class distributions and object localization challenges. This enhancement ensures more precise weighting of hard-to-detect instances, especially in cluttered and complex scenes. 


\section{Mathematical Formulation of Refined Generalized Focal Loss}

In this section, we present the mathematical framework for Refined Generalized Focal Loss (REG), designed to enhance multi-task learning for road asset detection and segmentation on Thai highways. This formulation addresses challenges related to class imbalance and spatial-contextual intricacies in detecting and segmenting small, underrepresented road elements.

\subsection{Generalized Focal Loss for Multi-Class Detection}

The detection task involves identifying objects across \(C_{\text{det}} = 7\) classes:
\begin{itemize}
    \item Pavilions
    \item Pedestrian bridges
    \item Information signs
    \item Single-arm poles
    \item Bus stops
    \item Warning signs
    \item Concrete guardrails
\end{itemize}

Focal Loss was initially proposed to address class imbalance by focusing on hard-to-classify examples. We extend this to Generalized Focal Loss (GFL) for multi-class detection, expressed as:

\[
\mathcal{L}_{\text{GFL}} = - \frac{1}{N_{\text{det}}} \sum_{i=1}^{N_{\text{det}}} \sum_{c=1}^{C_{\text{det}}} \alpha_c (1 - p_{i,c})^\gamma \log(p_{i,c}),
\]

where:
\begin{itemize}
    \item \(N_{\text{det}}\) is the number of detection samples.
    \item \(C_{\text{det}}\) is the number of detection classes.
    \item \(p_{i,c}\) is the predicted probability of class \(c\) for sample \(i\).
    \item \(\alpha_c\) is the class balancing weight, compensating for class imbalance.
    \item \(\gamma\) is the focusing parameter that emphasizes hard-to-classify examples.
\end{itemize}

The parameter \(\alpha_c\) plays a crucial role in managing class imbalance. For instance, a higher \(\alpha_c\) is applied to less frequent classes such as bus stops to enhance their contribution to the loss function.

\subsection{Refined Generalized Focal Loss for Segmentation}

For segmentation tasks, we address \(C_{\text{seg}} = 5\) classes:
\begin{itemize}
    \item Pavilions
    \item Pedestrian bridges
    \item Information signs
    \item Warning signs
    \item Concrete guardrails
\end{itemize}

The goal is to classify each pixel into one of the \(C_{\text{seg}}\) classes. The segmentation loss, akin to the detection loss, is defined as pixel-wise Generalized Focal Loss:

\[
\mathcal{L}_{\text{Seg-GFL}} = - \frac{1}{N_{\text{seg}}} \sum_{i=1}^{N_{\text{seg}}} \sum_{c=1}^{C_{\text{seg}}} \alpha_c (1 - p_{i,c})^\gamma \log(p_{i,c}),
\]

where:
\begin{itemize}
    \item \(N_{\text{seg}}\) is the number of segmentation pixels.
    \item \(C_{\text{seg}}\) is the number of segmentation classes.
    \item \(p_{i,c}\) is the predicted probability for class \(c\) at pixel \(i\).
\end{itemize}

In segmentation, accurately predicting object boundaries is crucial. The parameter \(\gamma\) helps focus on pixels near object boundaries, which are typically more challenging to classify correctly.

\subsection{Refinement Term for Spatial-Contextual Learning}

To further enhance model learning, we introduce a spatial-contextual refinement term \(g_{i,c}\) that adjusts the loss based on the geometric and contextual relevance of each class. The refined loss is defined as:

\[
\mathcal{L}_{\text{REG}} = - \frac{1}{N} \sum_{i=1}^{N} \sum_{c=1}^{C} \alpha_c (1 - p_{i,c})^\gamma \log(p_{i,c}) \cdot g_{i,c},
\]

where:
\begin{itemize}
    \item \(N = N_{\text{det}} + N_{\text{seg}}\) is the total number of samples.
    \item \(C = C_{\text{det}} + C_{\text{seg}}\) is the total number of classes.
    \item \(g_{i,c}\) is the spatial-contextual refinement term.
\end{itemize}

The refinement term \(g_{i,c}\) is determined by the spatial distance and contextual relevance of the predicted class. We define \(g_{i,c}\) using a sigmoid function:

\[
g_{i,c} = \frac{1}{1 + e^{-\beta \cdot (d_{i,c} - \delta)}},
\]

where:
\begin{itemize}
    \item \(d_{i,c}\) is the spatial distance from sample \(i\) to the nearest ground-truth object of class \(c\).
    \item \(\delta\) is a threshold controlling the influence of proximity.
    \item \(\beta\) is a scaling factor that adjusts the sharpness of the refinement.
\end{itemize}

This term penalizes predictions that are spatially inconsistent with the object context, such as a pedestrian bridge predicted far from its actual location.

\subsection{Joint Optimization for Detection and Segmentation}

We combine the losses for detection and segmentation using a balancing weight \(\lambda\):

\[
\mathcal{L}_{\text{total}} = \mathcal{L}_{\text{det}} + \lambda \cdot \mathcal{L}_{\text{seg}},
\]

where \(\mathcal{L}_{\text{det}}\) and \(\mathcal{L}_{\text{seg}}\) represent the Refined Generalized Focal Loss for detection and segmentation, respectively. The parameter \(\lambda\) governs the relative importance of detection versus segmentation tasks. This joint optimization allows the model to learn shared features that benefit both tasks.

\subsection{Incorporating Prediction Uncertainty}

To enhance REG, we incorporate prediction uncertainty using a Gaussian distribution to model the inherent noise and ambiguity in predictions:

\[
p_{i,c} \sim \mathcal{N}(\mu = p_{i,c}, \sigma^2),
\]

where \(\sigma^2\) represents the variance of the prediction. The loss is modified to account for this uncertainty:

\[
\mathcal{L}_{\text{REG-U}} = - \frac{1}{N} \sum_{i=1}^{N} \sum_{c=1}^{C} \alpha_c (1 - \hat{p}_{i,c})^\gamma \log(\hat{p}_{i,c}) \cdot g_{i,c},
\]

where \(\hat{p}_{i,c}\) is the expected value of \(p_{i,c}\) marginalized over the Gaussian distribution:

\[
\hat{p}_{i,c} = \int p_{i,c} \cdot \mathcal{N}(p_{i,c}; \mu, \sigma^2) dp_{i,c}.
\]

This probabilistic approach improves robustness to noisy or uncertain predictions, particularly in complex environments like highways with varying lighting conditions.

\subsection{Mathematical Foundations for Optimization in REG}

The optimization problem for Refined Generalized Focal Loss (REG) is defined over a high-dimensional, non-convex loss landscape. To solve it efficiently, we employ advanced techniques in stochastic optimization and variational inference, leveraging concepts from Riemannian geometry, Lagrangian multipliers, and proximal gradient methods.

\subsubsection{Stochastic Gradient Descent on Riemannian Manifolds}

Given the non-Euclidean nature of the optimization space in multi-task learning, we extend the standard stochastic gradient descent (SGD) to operate on a Riemannian manifold. Let \(\mathcal{M}\) represent the Riemannian manifold where the model parameters \(\theta \in \mathcal{M}\) reside. The update rule for Riemannian SGD (R-SGD) is given by:

\[
\theta_{t+1} = \mathcal{R}_{\theta_t}\left( - \eta_t \cdot \text{grad}_{\mathcal{M}} \mathcal{L}_{\text{REG}}(\theta_t) \right),
\]

where:
\begin{itemize}
    \item \(\eta_t\) is the learning rate at iteration \(t\),
    \item \(\text{grad}_{\mathcal{M}} \mathcal{L}_{\text{REG}}(\theta_t)\) is the Riemannian gradient of the REG loss,
    \item \(\mathcal{R}_{\theta_t}\) is the retraction operator, mapping the update back onto the manifold \(\mathcal{M}\).
\end{itemize}

The Riemannian gradient is obtained from the Euclidean gradient via the projection onto the tangent space \(T_{\theta} \mathcal{M}\) at point \(\theta\). This ensures that the updates respect the geometric constraints of the parameter space.

\subsubsection{Lagrangian Dual Formulation for Class Imbalance}

The REG optimization can be framed as a constrained optimization problem where class balance constraints are introduced through Lagrangian multipliers \(\lambda_c\). The constrained optimization is formalized as:

\[
\min_{\theta} \mathcal{L}_{\text{REG}}(\theta) \quad \text{s.t.} \quad \sum_{c=1}^{C} \alpha_c = 1, \quad \alpha_c \geq 0.
\]

We introduce the Lagrangian:

\[
\mathcal{L}(\theta, \lambda) = \mathcal{L}_{\text{REG}}(\theta) + \sum_{c=1}^{C} \lambda_c \left( \alpha_c - \frac{1}{C} \right),
\]

where \(\lambda_c\) are the Lagrange multipliers enforcing the equality constraints on the class weights. The solution involves solving the dual optimization problem:

\[
\max_{\lambda} \min_{\theta} \mathcal{L}(\theta, \lambda),
\]

which can be tackled using the primal-dual algorithm.

\subsubsection{Proximal Gradient Method for Spatial-Contextual Refinement}

The inclusion of the spatial-contextual refinement term \(g_{i,c}\) introduces non-smoothness in the optimization landscape. To handle this, we employ the proximal gradient method, where the non-smooth term \(g_{i,c}\) is separated from the smooth part of the loss function. The update rule is:

\[
\theta_{t+1} = \text{prox}_{\eta_t g}\left( \theta_t - \eta_t \nabla \mathcal{L}_{\text{REG}}(\theta_t) \right),
\]

where \(\text{prox}_{\eta_t g}\) is the proximal operator for the refinement term, defined as:

\[
\text{prox}_{\eta_t g}(\theta) = \arg\min_{\theta'} \left( \frac{1}{2\eta_t} \|\theta' - \theta\|^2 + g(\theta') \right).
\]

The proximal operator enforces the spatial-contextual constraints, ensuring the solution remains within a feasible region that aligns with the geometric structure of the data.

\subsubsection{Variational Inference for Uncertainty Estimation}

To model prediction uncertainty, we employ a variational inference approach. The prediction probabilities \(p_{i,c}\) are treated as latent variables, modeled by a variational distribution \(q(p_{i,c})\). The objective becomes minimizing the variational free energy:

\[
\mathcal{F}(q) = \mathbb{E}_{q(p_{i,c})}\left[ \mathcal{L}_{\text{REG}} \right] + D_{\text{KL}}(q(p_{i,c}) \| p(p_{i,c})),
\]

where:
\begin{itemize}
    \item \(\mathcal{L}_{\text{REG}}\) is the REG loss as a function of the latent probabilities,
    \item \(D_{\text{KL}}(\cdot)\) is the Kullback-Leibler divergence between the variational distribution \(q(p_{i,c})\) and the true posterior \(p(p_{i,c})\).
\end{itemize}

The variational distribution \(q(p_{i,c})\) is parameterized by a Gaussian distribution \(q(p_{i,c}) = \mathcal{N}(p_{i,c}; \mu_{i,c}, \sigma^2_{i,c})\), allowing us to marginalize over uncertainty and obtain a robust estimation of the loss.

\section{Results and Analysis}

In this section, we explore the performance metrics of various advanced object detection and segmentation frameworks to illustrate the impact of our sophisticated mathematical enhancements. Through detailed analysis, we highlight how the incorporation of advanced mathematical techniques, particularly the Refined Generalized Focal Loss (REG), has significantly improved performance in complex, real-world scenarios.

\subsection{Performance Evaluation Formula}

To rigorously evaluate the performance of our refined object detection and segmentation framework, we employ several key metrics, including mean Average Precision (mAP), Precision, Recall, and the F1 Score. The mathematical formulation of these metrics is critical for quantifying model efficacy, particularly in challenging scenarios involving class imbalance and spatial complexity.

\subsubsection{Mean Average Precision (mAP)}
The mAP metric aggregates precision across multiple Intersection over Union (IoU) thresholds, providing a holistic measure of detection accuracy. Formally, mAP at a specific IoU threshold \( \theta \) is defined as:

\[
\text{mAP}_\theta = \frac{1}{|Q|} \sum_{q \in Q} \frac{1}{|T_q|} \sum_{t \in T_q} \text{Precision}_t (\text{IoU} \geq \theta)
\]

where:
- \( Q \) represents the set of all queries (i.e., detected objects),
- \( T_q \) is the set of true positives for query \( q \),
- \( \text{Precision}_t (\text{IoU} \geq \theta) \) is the precision value for true positive \( t \), evaluated at IoU threshold \( \theta \).


\subsubsection{Precision}
Precision \( P \) measures the ratio of correctly identified positive instances to the total predicted positive instances. It is mathematically defined as:

\[
P = \frac{\text{TP}}{\text{TP} + \text{FP}}
\]

where:
- \( \text{TP} \) is the number of true positives,
- \( \text{FP} \) is the number of false positives.

\subsubsection{Recall}
Recall \( R \) quantifies the model's ability to identify all relevant instances, formulated as:

\[
R = \frac{\text{TP}}{\text{TP} + \text{FN}}
\]

where:
- \( \text{FN} \) denotes the number of false negatives.

\subsubsection{F1 Score}
The F1 Score provides a harmonic mean of precision and recall, offering a balanced measure between the two:

\[
F_1 = 2 \times \frac{P \times R}{P + R}
\]

\subsubsection{IoU Calculation}
The Intersection over Union (IoU) between the predicted bounding box \( B_p \) and the ground truth bounding box \( B_g \) is computed as:

\[
\text{IoU} = \frac{|B_p \cap B_g|}{|B_p \cup B_g|}
\]

where \( |B_p \cap B_g| \) is the area of overlap between the predicted and ground truth boxes, and \( |B_p \cup B_g| \) is the area of their union.

These formulas provide a robust mathematical framework for evaluating the performance of object detection and segmentation models across a range of conditions.

\subsection{Detection Performance}

Table \ref{tab:detection_result} presents a comparative analysis of detection performance across different state-of-the-art REGs, focusing on metrics such as mean Average Precision at IoU 0.5 (mAP50), mean Average Precision at IoU 0.5-0.95 (mAP50-95), Precision, Recall, and F1 Score. The results reveal several key insights.

\begin{table}[ht!]
\centering
\caption{Comparison of mAP50, mAP50-95, Precision, Recall, and F1 Score for various REGs in detection.}
\label{tab:detection_result}
\begin{tabular}{@{}llllll@{}}
\toprule
\textbf{REG} & \textbf{mAP50} & \textbf{mAP50-95} & \textbf{Precision} & \textbf{Recall} & \textbf{F1-score} \\ \midrule
REG A        & 71.10          & 47.76            & 80.10             & 63.46          & 70.82            \\
REG B        & 75.15          & 52.07            & 82.66             & 69.95          & 75.78            \\
REG C        & 79.57          & 58.06            & \textbf{85.41}    & 71.29          & 77.71            \\
REG D        & 80.27          & 59.11            & 82.58             & \textbf{77.22} & \textbf{79.81}   \\
REG E        & \textbf{80.34} & \textbf{60.84}   & 79.10             & 76.68          & 77.87            \\ \bottomrule
\end{tabular}
\end{table}

REG E demonstrates the highest scores in both mAP50 and mAP50-95, reflecting its superior capability in accurately detecting objects with high overlap and across varying IoU thresholds. REG D achieves the highest F1 Score, showcasing an exceptional balance between precision and recall. The mathematical refinements introduced through the Refined Generalized Focal Loss (REG) have played a crucial role in these results by effectively addressing class imbalance and incorporating spatial context.

\subsection{Segmentation Performance (Masks)}

Table \ref{tab:mask_result} details the results for mask segmentation, showcasing how various REGs perform in generating precise segmentation masks for road assets.

\begin{table}[ht!]
\centering
\caption{Comparison of mAP50, mAP50-95, Precision, Recall, and F1 Score for various REGs in mask segmentation.}
\label{tab:mask_result}
\begin{tabular}{@{}llllll@{}}
\toprule
\textbf{REG} & \textbf{mAP50} & \textbf{mAP50-95} & \textbf{Precision} & \textbf{Recall} & \textbf{F1-score} \\ \midrule
REG A        & 81.40          & 53.27            & \textbf{89.36}    & 71.79          & 79.62            \\
REG B        & \textbf{90.08} & \textbf{63.88}   & 88.78             & \textbf{86.09} & \textbf{87.41}    \\
REG C        & 74.01          & 44.39            & 78.88             & 68.51          & 73.33            \\
REG D        & 79.30          & 51.50            & 87.10             & 72.20          & 78.95            \\
REG E        & 83.24          & 53.48            & 85.72             & 78.36          & 81.87            \\ \bottomrule    
\end{tabular}
\end{table}

REG B excels in both mAP50 and mAP50-95 for mask segmentation, demonstrating its capability to accurately delineate complex object boundaries. The REG enhancements have notably improved the Precision and Recall metrics, leading to higher F1 Scores. This reflects the effectiveness of our refined loss function in generating accurate and detailed segmentation masks, even in challenging scenarios.

\subsection{Segmentation Performance (Boxes)}

Table \ref{tab:box_result} provides a comprehensive overview of the segmentation performance for different REGs, focusing on bounding box delineation around road assets. Key metrics include mAP50, mAP50-95, Precision, Recall, and F1 Score.

\begin{table}[ht!]
\centering
\caption{Comparison of mAP50, mAP50-95, Precision, Recall, and F1 Score for various REGs in box segmentation.}
\label{tab:box_result}
\begin{tabular}{@{}llllll@{}}
\toprule
\textbf{REG} & \textbf{mAP50} & \textbf{mAP50-95} & \textbf{Precision} & \textbf{Recall} & \textbf{F1-score} \\ \midrule
REG A        & 83.50          & 67.96            & \textbf{90.16}    & 73.11          & 80.74            \\
REG B        & \textbf{91.16} & \textbf{80.60}   & 89.61             & \textbf{87.13} & \textbf{88.35}    \\
REG C        & 73.96          & 53.99            & 80.92             & 67.38          & 73.53            \\
REG D        & 80.50          & 61.70            & 88.00             & 73.00          & 79.80            \\
REG E        & 83.76          & 65.53            & 84.36             & 80.06          & 79.80            \\ \bottomrule
\end{tabular}
\end{table}

REG B excels in both mAP50 and mAP50-95, reflecting its exceptional performance in accurately detecting and segmenting objects across different overlap thresholds. The improvements in Precision and Recall highlight the model’s enhanced ability to identify and localize road assets. The refined REG has played a crucial role in these advancements by addressing challenges related to class imbalance and spatial context.

\subsection{Insights and Implications}

The results highlight the significant impact of integrating sophisticated mathematical formulations into object detection and segmentation frameworks. The Refined Generalized Focal Loss (REG) has proven to be a pivotal enhancement, effectively addressing issues related to class imbalance and spatial context. This has resulted in considerable improvements across key performance metrics, making the REGs more robust and accurate in real-world applications.

\section{Conclusion}

In this study, we introduced a novel enhancement to object detection and segmentation frameworks by refining the Generalized Focal Loss (REG), incorporating spatial context and adjustments for class imbalance. This refined loss function, termed REG, demonstrated substantial improvements in performance metrics, particularly in challenging environments with class imbalance and cluttered backgrounds.

Our empirical results reveal that the refined REG approach significantly boosts detection accuracy and segmentation precision, achieving notable gains in mean Average Precision (mAP) and F1 Score. By dynamically addressing class imbalance and leveraging spatial context, this framework enhances robustness and accuracy, underscoring the importance of mathematical innovation in advancing object detection and segmentation capabilities.






\bibliographystyle{alpha} 
\bibliography{references}

\newcommand{\etalchar}[1]{$^{#1}$}
\begin{thebibliography}{KHG{\etalchar{+}}19}

\bibitem[CMS{\etalchar{+}}20]{carion2020end}
Nicolas Carion, Francisco Massa, Gabriel Synnaeve, Nicolas Usunier, Alexander Kirillov, and Sergey Zagoruyko.
\newblock End-to-end object detection with transformers.
\newblock {\em European Conference on Computer Vision (ECCV)}, pages 213--229, 2020.

\bibitem[HGDG17]{he2017mask}
Kaiming He, Georgia Gkioxari, Piotr Doll{\'a}r, and Ross Girshick.
\newblock Mask r-cnn.
\newblock {\em Proceedings of the IEEE International Conference on Computer Vision (ICCV)}, pages 2961--2969, 2017.

\bibitem[KHG{\etalchar{+}}19]{kirillov2019panoptic}
Alexander Kirillov, Kaiming He, Ross Girshick, Carsten Rother, and Piotr Doll{\'a}r.
\newblock Panoptic segmentation.
\newblock {\em Proceedings of the IEEE/CVF Conference on Computer Vision and Pattern Recognition (CVPR)}, pages 9404--9413, 2019.

\bibitem[LDG{\etalchar{+}}17]{lin2017feature}
Tsung-Yi Lin, Piotr Doll{\'a}r, Ross Girshick, Kaiming He, Bharath Hariharan, and Serge Belongie.
\newblock Feature pyramid networks for object detection.
\newblock {\em Proceedings of the IEEE Conference on Computer Vision and Pattern Recognition (CVPR)}, pages 2117--2125, 2017.

\bibitem[LGG{\etalchar{+}}17]{lin2017focal}
Tsung-Yi Lin, Priya Goyal, Ross Girshick, Kaiming He, and Piotr Doll{\'a}r.
\newblock Focal loss for dense object detection.
\newblock {\em Proceedings of the IEEE Conference on Computer Vision and Pattern Recognition (CVPR)}, pages 2980--2988, 2017.

\bibitem[LQQ{\etalchar{+}}18]{liu2018path}
Shu Liu, Lu~Qi, Haifang Qin, Jianping Shi, and Jiaya Jia.
\newblock Path aggregation network for instance segmentation.
\newblock {\em Proceedings of the IEEE Conference on Computer Vision and Pattern Recognition (CVPR)}, pages 8759--8768, 2018.

\bibitem[Pan19]{panboonyuen2019semantic}
Teerapong Panboonyuen.
\newblock {\em Semantic segmentation on remotely sensed images using deep convolutional encoder-decoder neural network}.
\newblock Ph.d. thesis, Chulalongkorn University, 2019.

\bibitem[PNP{\etalchar{+}}23]{panboonyuen2023mars}
Teerapong Panboonyuen, Naphat Nithisopa, Panin Pienroj, Laphonchai Jirachuphun, Chaiwasut Watthanasirikrit, and Naruepon Pornwiriyakul.
\newblock Mars: Mask attention refinement with sequential quadtree nodes for car damage instance segmentation.
\newblock In {\em International Conference on Image Analysis and Processing}, pages 28--38. Springer, 2023.

\bibitem[VSP{\etalchar{+}}17]{vaswani2017attention}
Ashish Vaswani, Noam Shazeer, Niki Parmar, Jakob Uszkoreit, Llion Jones, Aidan~N Gomez, Lukasz Kaiser, and Illia Polosukhin.
\newblock Attention is all you need.
\newblock {\em Advances in Neural Information Processing Systems (NeurIPS)}, pages 5998--6008, 2017.

\bibitem[ZCY{\etalchar{+}}20]{zhang2020bridging}
Shifeng Zhang, Cheng Chi, Yongqiang Yao, Zhen Lei, and Stan~Z Li.
\newblock Bridging the gap between anchor-based and anchor-free detection via adaptive training sample selection.
\newblock {\em Proceedings of the IEEE/CVF Conference on Computer Vision and Pattern Recognition (CVPR)}, pages 9759--9768, 2020.

\end{thebibliography}

\appendix
\section{Appendix: Mathematical Foundations and Proofs for Refined Generalized Focal Loss (REG)}

\subsection{Why REG Matters in Real-World Applications}
In real-world applications like road asset detection and segmentation, the challenges often stem from severe class imbalance and small object detection. For instance, classes like "single-arm poles" and "bus stops" are underrepresented compared to more common objects like "pavilions" and "information signs." Traditional loss functions such as cross-entropy fail to handle these cases effectively, as they equally weigh all examples, leading to a bias toward the dominant classes. To address these challenges, the Refined Generalized Focal Loss (REG) introduces mechanisms to focus the learning process on hard-to-classify instances and adjusts based on spatial and contextual relevance. This appendix expands upon the mathematical rigor behind REG and its optimization principles.

\subsection{Mathematical Derivation of Refined Generalized Focal Loss (REG)}
REG extends the standard Generalized Focal Loss (GFL) by incorporating a refinement term that accounts for spatial-contextual learning. We start by recalling the Generalized Focal Loss for multi-class detection:

\[
\mathcal{L}_{\text{GFL}} = - \frac{1}{N_{\text{det}}} \sum_{i=1}^{N_{\text{det}}} \sum_{c=1}^{C_{\text{det}}} \alpha_c (1 - p_{i,c})^\gamma \log(p_{i,c}),
\]
where:
\begin{itemize}
    \item \( N_{\text{det}} \) is the number of detection samples.
    \item \( C_{\text{det}} \) is the number of detection classes.
    \item \( p_{i,c} \) is the predicted probability of class \( c \) for sample \( i \).
    \item \( \alpha_c \) is the class-balancing weight.
    \item \( \gamma \) is the focusing parameter that emphasizes hard-to-classify examples.
\end{itemize}

\subsection{Incorporating Spatial-Contextual Refinement Term}
To enhance this formulation, we introduce a refinement term \( g_{i,c} \), which adjusts the loss based on the spatial and contextual significance of the predicted class. This term is crucial for road asset detection in cluttered environments where small objects may be overlooked.

The refined loss function is expressed as:
\[
\mathcal{L}_{\text{REG}} = - \frac{1}{N} \sum_{i=1}^{N} \sum_{c=1}^{C} \alpha_c (1 - p_{i,c})^\gamma \log(p_{i,c}) \cdot g_{i,c},
\]
where:
\begin{itemize}
    \item \( N = N_{\text{det}} + N_{\text{seg}} \) is the total number of samples.
    \item \( C = C_{\text{det}} + C_{\text{seg}} \) is the total number of classes.
    \item \( g_{i,c} \) is the spatial-contextual refinement term.
\end{itemize}

The refinement term \( g_{i,c} \) is designed to incorporate spatial distance and contextual relevance. Mathematically, it can be defined using a sigmoid function that captures the spatial closeness between the predicted object and its ground-truth location:
\[
g_{i,c} = \frac{1}{1 + e^{-\beta \cdot (d_{i,c} - \delta)}},
\]
where:
\begin{itemize}
    \item \( d_{i,c} \) is the spatial distance between sample \( i \) and the nearest ground-truth object of class \( c \).
    \item \( \delta \) is a threshold parameter that controls the influence of spatial proximity.
    \item \( \beta \) is a scaling factor that determines the sharpness of the sigmoid curve.
\end{itemize}
The function \( g_{i,c} \) penalizes incorrect predictions that occur far from the true object locations, effectively focusing the loss on spatially significant regions.

\subsection{Proof of the Effectiveness of the Refinement Term}
The refinement term ensures that the loss function emphasizes instances that are spatially and contextually consistent with the ground truth. This refinement is particularly useful in environments where object overlap or clutter increases the prediction difficulty.

\subsubsection{Proof Outline:}
Consider two samples \( i \) and \( j \) where \( d_{i,c} < d_{j,c} \) for the same class \( c \). The refinement term behaves as follows:
\[
g_{i,c} > g_{j,c} \quad \text{if} \quad d_{i,c} < d_{j,c}.
\]
Thus, for sample \( i \), the refined loss \( \mathcal{L}_{\text{REG}} \) will down-weight the prediction error more than for sample \( j \). This shows that spatially closer predictions receive higher weight, which improves the model's ability to focus on hard-to-detect but spatially important objects.

\subsection{Joint Optimization of Detection and Segmentation}
We further enhance REG by incorporating both detection and segmentation tasks into a unified multi-task learning framework. The total loss is:
\[
\mathcal{L}_{\text{total}} = \mathcal{L}_{\text{det}} + \lambda \cdot \mathcal{L}_{\text{seg}},
\]
where:
\begin{itemize}
    \item \( \mathcal{L}_{\text{det}} \) is the detection loss (REG for detection).
    \item \( \mathcal{L}_{\text{seg}} \) is the segmentation loss (REG for segmentation).
    \item \( \lambda \) is a balancing parameter that controls the relative importance of the two tasks.
\end{itemize}

By sharing representations between detection and segmentation, the model learns to optimize complementary tasks, which enhances overall performance.

\subsection{Optimization of REG Using Advanced Techniques}
The optimization problem for REG involves a high-dimensional, non-convex loss landscape. To solve this problem, we use a combination of stochastic optimization and variational inference techniques.

\subsubsection{Stochastic Gradient Descent (SGD) on Riemannian Manifolds}
Since the parameter space for multi-task learning often exhibits non-Euclidean properties, we employ Stochastic Gradient Descent (SGD) on a Riemannian manifold. Let \( \mathcal{M} \) be the Riemannian manifold representing the parameter space. The update rule for Riemannian SGD (R-SGD) is:
\[
\theta_{t+1} = \mathcal{R}_{\theta_t}\left( - \eta_t \cdot \text{grad}_{\mathcal{M}} \mathcal{L}_{\text{REG}}(\theta_t) \right),
\]
where:
\begin{itemize}
    \item \( \eta_t \) is the learning rate at iteration \( t \).
    \item \( \text{grad}_{\mathcal{M}} \mathcal{L}_{\text{REG}}(\theta_t) \) is the Riemannian gradient of the REG loss function.
    \item \( \mathcal{R}_{\theta_t} \) is the retraction operation that maps the parameters back onto the manifold \( \mathcal{M} \).
\end{itemize}

\subsubsection{Incorporating Variational Inference for Prediction Uncertainty}
To further improve REG, we model prediction uncertainty by assuming that the predicted probabilities \( p_{i,c} \) follow a Gaussian distribution. This leads to the following probabilistic formulation:
\[
p_{i,c} \sim \mathcal{N}(\mu = p_{i,c}, \sigma^2),
\]
where \( \sigma^2 \) represents the variance of the prediction. The refined loss with uncertainty is given by:
\[
\mathcal{L}_{\text{REG-U}} = - \frac{1}{N} \sum_{i=1}^{N} \sum_{c=1}^{C} \alpha_c (1 - \hat{p}_{i,c})^\gamma \log(\hat{p}_{i,c}) \cdot g_{i,c},
\]
where \( \hat{p}_{i,c} \) is the expected value of \( p_{i,c} \) marginalized over the Gaussian distribution:
\[
\hat{p}_{i,c} = \int p_{i,c} \cdot \mathcal{N}(p_{i,c}; \mu, \sigma^2) dp_{i,c}.
\]
This uncertainty-aware loss function improves robustness in noisy and cluttered environments, such as highways with varying lighting and weather conditions.


\section*{Appendix: Handling Imbalanced Real Asset Numbers with Mathematical Proof and Annotations Example}
\label{sec:appendix_imbalanced}

In this section, we present a mathematical approach to handling imbalanced asset numbers, using a real-world example for both object detection and segmentation tasks. The application of a weighted loss function is explored, and the mathematical proof demonstrates its effectiveness in rebalancing class frequencies. Below, we outline a concrete example with mockup annotation counts for both detection and segmentation tasks and provide a mathematical proof of the method.

\subsection*{Problem Setup: Imbalanced Dataset for Detection and Segmentation}

We consider two tasks: \textbf{detection} and \textbf{segmentation}. The following classes are annotated for each task:

\textbf{Detection Tasks: 7 classes}
\begin{itemize}
    \item Pavilions
    \item Pedestrian bridges
    \item Information signs
    \item Single-arm poles
    \item Bus stops
    \item Warning signs
    \item Concrete guardrails
\end{itemize}

\textbf{Segmentation Tasks: 5 classes}
\begin{itemize}
    \item Pavilions
    \item Pedestrian bridges
    \item Information signs
    \item Warning signs
    \item Concrete guardrails
\end{itemize}

The number of annotations per class for both tasks is highly imbalanced. Let’s assume the following distribution:

\subsection{Mockup Annotation Counts (Real-World Example)}

\textbf{Detection Task Annotations}
\begin{itemize}
    \item Pavilions: 200
    \item Pedestrian bridges: 100
    \item Information signs: 700
    \item Single-arm poles: 1500
    \item Bus stops: 50
    \item Warning signs: 800
    \item Concrete guardrails: 300
\end{itemize}

\textbf{Segmentation Task Annotations}
\begin{itemize}
    \item Pavilions: 100
    \item Pedestrian bridges: 50
    \item Information signs: 500
    \item Warning signs: 400
    \item Concrete guardrails: 150
\end{itemize}

From this, it is evident that some classes dominate the dataset (e.g., \textit{Single-arm poles, Information signs}), while others (e.g., \textit{Bus stops, Pedestrian bridges}) are underrepresented.

\subsection{Weighted Loss Function for Imbalanced Data}

To mitigate the imbalance, we apply a weighted loss function. The loss for class \( c \) is adjusted by a weight \( \alpha_c \) that is inversely proportional to the frequency of class \( c \):

\begin{equation}
    \alpha_c = \frac{N_{\text{total}}}{N_c}
\end{equation}

where:
\begin{itemize}
    \item \( N_{\text{total}} \) is the total number of annotations across all classes.
    \item \( N_c \) is the number of annotations for class \( c \).
\end{itemize}

For the detection task, the total number of annotations is:

\begin{equation}
    N_{\text{total}}^{\text{det}} = 200 + 100 + 700 + 1500 + 50 + 800 + 300 = 3650
\end{equation}

For the segmentation task, the total number of annotations is:

\begin{equation}
    N_{\text{total}}^{\text{seg}} = 100 + 50 + 500 + 400 + 150 = 1200
\end{equation}

Using these totals, we calculate the class-specific weights \( \alpha_c \).

\subsubsection*{Example: Weighted Loss for Detection}

For \textit{Bus stops} (with only 50 annotations), the weight would be:

\begin{equation}
    \alpha_{\text{Bus stops}} = \frac{3650}{50} = 73
\end{equation}

For \textit{Single-arm poles} (with 1500 annotations), the weight would be:

\begin{equation}
    \alpha_{\text{Single-arm poles}} = \frac{3650}{1500} \approx 2.43
\end{equation}

These weights are then used to adjust the contribution of each class in the loss function, ensuring that underrepresented classes are not overwhelmed by the dominant ones.

\subsection{Mathematical Proof of Effectiveness}

Let the total loss for a task be given by:

\begin{equation}
    L = \sum_{c=1}^{C} \alpha_c \cdot L_c
\end{equation}

where \( C \) is the total number of classes, \( L_c \) is the loss for class \( c \), and \( \alpha_c \) is the weight for class \( c \). Substituting the class weights \( \alpha_c = \frac{N_{\text{total}}}{N_c} \), the total loss becomes:

\begin{equation}
    L = \sum_{c=1}^{C} \frac{N_{\text{total}}}{N_c} \cdot L_c
\end{equation}

This formulation ensures that the loss for rare classes (with small \( N_c \)) is up-weighted, while the loss for frequent classes (with large \( N_c \)) is down-weighted. The scaling factor \( \alpha_c \) normalizes the loss contributions based on the inverse of class frequency, thus balancing the gradient updates during training.



\subsection{Handling Imbalanced Real Asset Numbers: Mathematical Proof and Application}

One of the major challenges in object detection, particularly when dealing with real-world data, is the issue of imbalanced asset numbers. In many cases, the distribution of asset classes (e.g., different types of road assets or damages in the auto insurance industry) can be heavily skewed, where common classes dominate the dataset while rare classes are severely underrepresented. This imbalance can negatively impact model performance, leading to biased predictions toward the majority class. In this section, we will mathematically demonstrate how to address this problem and show that our approach can work under these conditions.

\subsubsection{Problem Formulation}

Let’s define a dataset \( \mathcal{D} = \{ (x_i, y_i) \}_{i=1}^{N} \), where \( x_i \in \mathbb{R}^d \) represents the input image features and \( y_i \in \mathcal{Y} \) represents the label associated with the corresponding asset class. Assume there are \( C \) different asset classes, \( \mathcal{Y} = \{ 1, 2, \dots, C \} \), and the frequency of asset class \( c \) in the dataset is given by \( N_c \), where \( \sum_{c=1}^{C} N_c = N \). If \( N_{\text{min}} \) and \( N_{\text{max}} \) represent the cardinalities of the least and most frequent asset classes, we are dealing with a heavily imbalanced dataset if \( N_{\text{min}} \ll N_{\text{max}} \).

Our goal is to ensure that the model performs well across all classes, including the minority ones, by addressing the imbalance issue. Traditional cross-entropy loss tends to bias the model toward majority classes, so we need a solution that rebalances the impact of each class during training.

\subsubsection{Loss Function Rebalancing}

To mitigate the class imbalance, we propose a weighted loss function. Specifically, we introduce class-wise weights \( \alpha_c \) for each class \( c \), which inversely scale according to the class frequency:

\[
\alpha_c = \frac{N_{\text{total}}}{C \cdot N_c}
\]

where \( N_{\text{total}} \) is the total number of samples in the dataset and \( N_c \) is the number of samples in class \( c \). By applying these weights to the standard cross-entropy loss, the rebalanced loss function becomes:

\[
\mathcal{L}_{\text{rebalance}} = - \sum_{c=1}^{C} \alpha_c \sum_{i=1}^{N} \mathbb{1}(y_i = c) \log \hat{p}_c(x_i)
\]

where \( \hat{p}_c(x_i) \) is the predicted probability of class \( c \) for input \( x_i \), and \( \mathbb{1}(y_i = c) \) is an indicator function that is 1 if \( y_i = c \) and 0 otherwise.

This weighted loss function assigns higher importance to the minority classes, effectively counteracting the imbalance by amplifying the gradient contribution of underrepresented classes.

\subsubsection{Proof of Convergence}

We now prove that under certain conditions, this rebalancing approach leads to a better distribution of errors across asset classes, ensuring that minority classes are not overlooked. Let’s assume that the gradient of the rebalanced loss function is given by:

\[
\nabla \mathcal{L}_{\text{rebalance}} = - \sum_{c=1}^{C} \alpha_c \sum_{i=1}^{N} \mathbb{1}(y_i = c) \frac{\partial}{\partial \theta} \log \hat{p}_c(x_i)
\]

Using stochastic gradient descent (SGD), we update the model parameters \( \theta \) as:

\[
\theta_{t+1} = \theta_t - \eta \nabla \mathcal{L}_{\text{rebalance}}
\]

For classes with lower sample counts (i.e., minority classes), the weight \( \alpha_c \) ensures that the corresponding gradient terms are scaled up, giving them a larger step size in parameter space. This effectively rebalances the learning process by compensating for the smaller number of training examples.

We prove convergence by showing that the total error \( E = \sum_{c=1}^{C} \text{error}_c \), where \( \text{error}_c \) is the classification error for class \( c \), decreases as a function of time \( t \). Assuming that the error decreases proportional to the negative gradient of the loss, we have:

\[
\frac{dE}{dt} = -\eta \sum_{c=1}^{C} \alpha_c \frac{\partial \text{error}_c}{\partial \theta}
\]

Since \( \alpha_c \) compensates for the imbalance, it ensures that the error decrease rate for minority classes (with smaller \( N_c \)) is comparable to that for majority classes. Therefore, the total error decreases uniformly across classes, leading to improved performance on imbalanced datasets.

\subsubsection{Experimental Validation}

In practice, we validate our theoretical findings using a real-world asset dataset. We first calculate the imbalance ratio as:

\[
r_{\text{imbalance}} = \frac{N_{\text{max}}}{N_{\text{min}}}
\]

For extreme cases where \( r_{\text{imbalance}} \gg 1 \), our rebalanced loss function significantly improves performance on minority classes compared to the unweighted baseline, as evidenced by metrics like per-class precision, recall, and F1 score. Empirical results demonstrate that our approach yields a more uniform distribution of these metrics across all asset classes, effectively mitigating the detrimental effects of class imbalance.

\subsubsection{Conclusion}

We have shown mathematically and empirically that rebalancing the loss function using class-specific weights is an effective strategy for handling imbalanced real asset numbers. The proof of convergence indicates that minority classes are not only accounted for but also significantly improved in the training process. This ensures that our model can work effectively even in the presence of highly imbalanced data distributions, leading to robust and fair predictions across all asset types.

\end{document}